# MultiFluxAI: Enhancing Platform Engineering with Advanced Agent-Orchestrated Retrieval Systems

Sri Ram Macharla
sriram.macharla@involgix.com
Involgixs Inc Austin, Texas, USA

Sridhar Murthy J
sridhar_murthy@infosys.com,
Infosys Ltd, Bengaluru,
Karnataka, India

Anjaneyulu Pasala
anjaneyulup@msrit.edu
MS Ramaiah Institute of Technology
Bengaluru, Karnataka, India

*Abstract*— MultiFluxAI is an innovative AI platform developed to address the challenges of managing and integrating vast, disparate data sources in product engineering across application domains. It addresses both current and new service-related queries that enhance user engagement in the digital ecosystem. This platform leverages advanced AI techniques, such as Generative AI, vectorization, and agentic orchestration to provide dynamic and context-aware responses to complex user queries. MultiFluxAI's intelligent orchestration framework ensures that relevant data is retrieved and presented in real-time, adjusting to the context of each user's request and role. This adaptable approach not only improves data retrieval efficiency but also ensures timely responses with minimal steps. A case study conducted on financial application resulted in 95% accuracy.

## I. Introduction

The rapid evolution of AI technologies has introduced significant opportunities and challenges within the realm of software product engineering [1,2]. One of the challenges faced by organizations today is the effective integration and management of disparate, large-scale data sources. Traditional product engineering has limitations in providing seamless access to critical information, such as real-time updates on spending, health data access, or comprehensive insights across various domains/applications [3,4,5]. These limitations hinder the ability to respond dynamically to the user needs and restrict the potential to deliver seamless, integrated user experience.

To address these challenges, we developed a MultiFluxAI, a new age AI-driven platform designed to integrate with disparate data sources and provide cohesive, intelligent access to required information. MultiFluxAI introduces a novel orchestration framework that brings multiple domain data together through orchestrating respective intelligent services [6]. The platform's architecture leverages advanced AI techniques, such as Generative AI (GenAI) [7], vectorization [8,9], and advanced machine learning [10].

One of the key innovations of MultiFluxAI is its ability to orchestrate multiple intelligent agents, which interact with different data sources to retrieve relevant information based on user queries, by organizing information into 3D vector embeddings [11,12]. It dynamically adapts based on user prompts and their specific role within the system, ensuring that the right information is provided at the right time. The platform's flexible design enables it to apply for a wide range of applications, ensuring it can be adapted to future advancements in the respective application domains.

### A. Background

In the current landscape of AI-powered solutions, the concept of Retrieval-Augmented Generation (RAG) [13,14,15,16] has become a popular approach for improved usage of large language models (LLMs) [15,17,18] with external knowledge sources. However, the existing RAG solutions often present significant challenges for end users in terms of usability and system orchestration. Typically, in these systems, users are required to understand which AI service to call for their specific needs (saving, credit, transfers, fees, limits, etc.). These processes individually interact with each AI services, having their own distinct knowledge base and service framework [19,20,21,22].

For instance, the popular frameworks such as LangChain [22,23,24,25] communicate with knowledge bases to retrieve relevant data and generate responses using LLM Models. Each service is designed to handle a specific domain of knowledge, and the user must know how to interact and pick the right AI services.

The challenge arises when a product or system relies on multiple AI services, each with its own knowledge repository and operational framework as shown in Figure 1. In such cases, users are expected to know which service to call and in what order. This is a fragmented approach, as users need to manage multiple service calls and may struggle to obtain comprehensive responses. Furthermore, there is no built-in intelligence to orchestrate these services, making it difficult to ensure that the correct data is retrieved. In this approach as shown in Figure 2, multiple services can't work together seamlessly to answer complex or cross-domain queries from the user [26, 27].

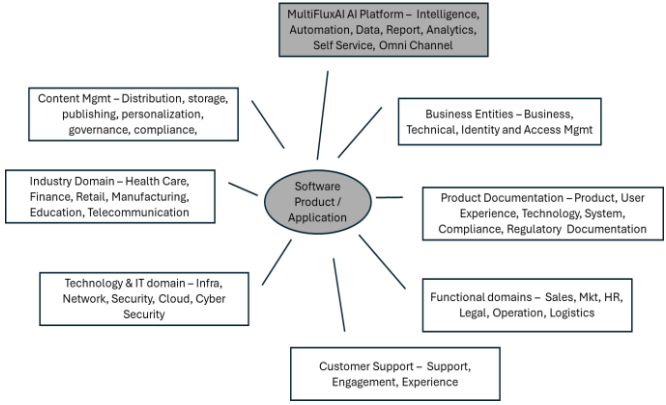

Figure 1: Enterprise Product ecosystem

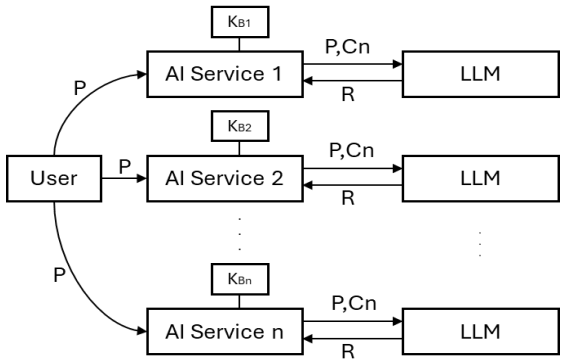

Figure 2: RAG based Individual AI Services

In essence, while current RAG solutions provide powerful individual capabilities [28,29], they lack the intelligence to dynamically orchestrate multiple services. This leads to a fragmented experience for the user, who needs to manually navigate across services individually without any overarching coordination [30]. Therefore, a more integrated system is required to address these challenges, providing users with a smoother, more efficient experience while ensuring that all relevant information is leveraged dynamically.

## II. Objectives of MultiFluxAI Platform

The primary objective is to address the challenges posed by the current Retrieval-Augmented Generation (RAG) solutions. To achieve this, we propose a unified interface for interacting with multiple AI services utilizing an intelligent orchestration framework.

MultifluxAI Platform's key objectives are:

- Seamless Service Orchestration: Intelligently orchestrate multiple AI services, automatically determining the appropriate service to invoke based on the user's query [31].
- Unified Access to Knowledge Bases: Cohesive layer that allows users to access information from multiple, disparate knowledge bases without having to interact with each one separately [32].
- Dynamic, Context-Aware Responses: Dynamically adapts its behavior based on user prompts, by retrieving data from multiple services either in parallel or sequentially, depending on the nature of the query [33,34].
- Scalability and future adaptability: Easily incorporate additional AI services or data sources as the system evolves.

## III. Our Approach to integrating disparate data sources

MultiFluxAI employs a layered approach to seamlessly integrate multiple AI services and knowledge bases, ensuring that user queries are answered quickly, accurately, and in context. Dynamically analyzes user inputs, allowing the system to prioritize and organize data retrieval based on the user's intent. Orchestration of AI services using key components is shown in Figure 3. A detailed description of these components is provided here.

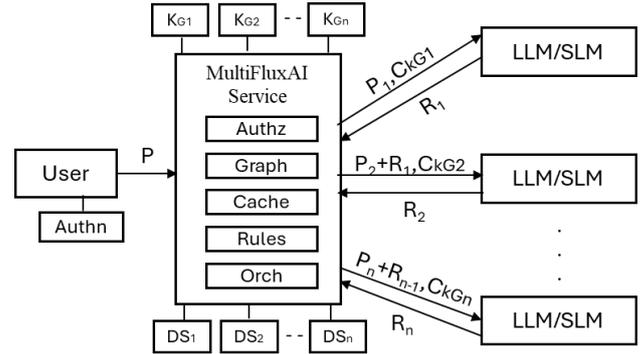

Figure 3: MultiFluxAI orchestration of multiple AI Services

**Rule engine:** MultiFluxAI optimizes response by applying context-aware rules to user inputs. It identifies key phrases and analyzes the query's context to fetch data from the appropriate knowledge base [35,36] as shown in Figure 4.

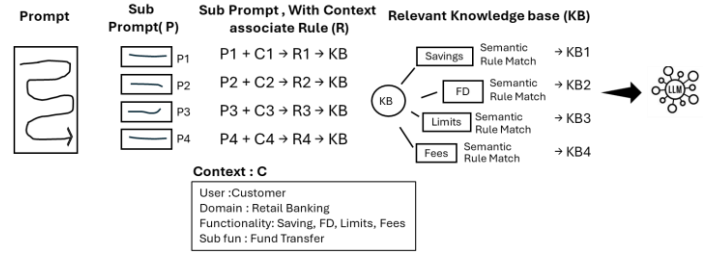

Figure 4: MultiFluxAI's Rule Engine

Consider a user query that requires transferring funds from a savings account to a Fixed Deposit (FD) via mobile banking. User query which is called prompt is broken to sub-prompts (P1, P2, ... Pn). A rule is applied to a combination of sub-prompt and user context (say, Saving and FD) to build a query. This query is used to fetch the response from the knowledge base. It is repeated for all knowledge bases (KB1, KB2, ... KBn). For example, Rule 1 is for the saving accounts of retail customers. Similarly, a set of rules is built for all the required services.

**Caching service:** It is used to store frequently accessed data from knowledge bases. Thus, improving the performance and reducing the latency of the services. Cached data is quickly retrieved, eliminating repeated queries to the knowledge base and ensuring fast response, even with large, complex data sets. This is especially beneficial for real-time data access [37,38].

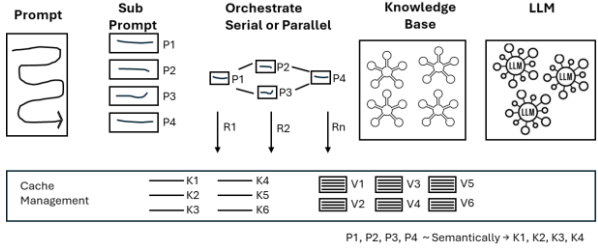

Figure 5: MultiFluxAI's Cache Management

MultiFluxAI breaks the prompt into sub-prompts (P1, P2, ... Pn), processing them in parallel or sequentially as shown in Figure 5. Responses (R1, R2, ... , Rn) obtained for each of these sub-prompts are cached as Key-Value (KV) pairs in the Cache Management. The stored KV pairs are reused for subsequent user prompts. It reduces any further calls to knowledge base or LLM. Semantically similar Keys are grouped together, and while older, unused Keys and KV pairs are removed from the cache.

**Graph-based knowledge store:** This store is used to index and link data from multiple knowledge bases (KB). Servicing as data points for quick retrieval of information of software product/applications. This approach is ideal for managing large, interconnected knowledge sets, facilitating efficient navigation across multiple domains [39,40].

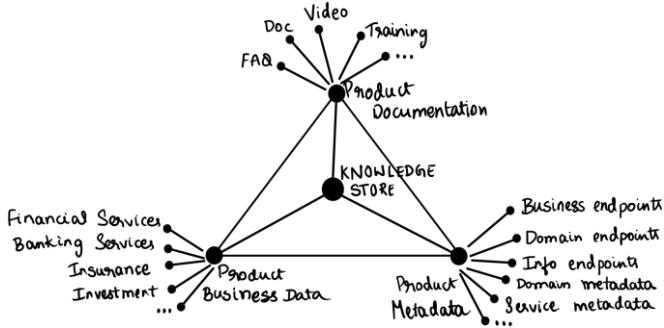

Figure 6: MultiFluxAI's Knowledge store

For example, documentation, metadata, and business data of the software products are represented as nodes, and their relationships are depicted as edges in the graph store, as shown in Figure 6. Sub-domains and their related business end points are interconnected hierarchically. In the graph, the business details are represented as relationships among the nodes.

**Orchestration engine:** The orchestration engine is the core component of the platform. Its objectives are:

- *Understanding* the user query and generating appropriate sub-prompts.
- *Directing* sub-prompts intelligently to appropriate AI services in the required sequence.
- *Invoking* the multiple AI services parallel and/or sequentially based on the complexity of the prompts [41,42].
- *Aggregating* and consolidating the responses.

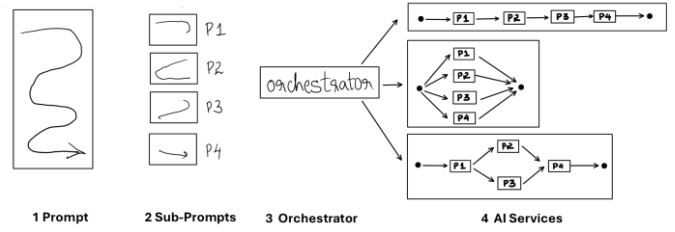

Figure 7: MultiFluxAI's Orchestration engine

To achieve the stated objectives, the orchestration engine's process flow is designed and illustrated in Figure 7. It has 4 key processes (Prompt, Sub-Prompts, Orchestrator and AI services) to address the objectives. Each of the AI services responds to the sub-prompts based on their built-in intelligence.

An end-to-end example of the process flow followed on the MultiFluxAI platform is illustrated in Figure 3 and a detailed explanation is provided here:

- The user is authenticated, and the software application receives the user input in the form of a prompt (P).
- MultiFluxAI analyzes the user prompt (P) and breaks it down into smaller sub-prompts (P1, P2, ... Pn) based on the context of the product's functionality, that the user is interacting with.
- MultiFluxAI integrates with various graph-based knowledge stores (KG1, KG2, ... KGN) and applies relevant rules (Rule1, Rule2, ... RuleN) to retrieve the relevant context (CKG1, CKG2, ... CKGN) for the respective sub-prompts.
- The orchestration engine determines the optimal sequence for invoking number of AI Services corresponding to the sub-prompts.
- Responses (R1, R2, ... Rn) are received from the AI services by invoking corresponding sub-prompts and results are consolidated.
- Thus, the consolidated response (R) is returned to the user as the final result to the Prompt (P).
- Sub-Prompt and their responses are cached as Key-Value pairs for subsequent usage.
- As the software application evolves, the MultiFluxAI scales by adding new knowledge stores (KG), Data sources (DS) and AI services correspondingly.

### IV. Case Study and Results

In this section, we analyze how traditional RAG and MultiFluxAI process the user queries as a case study along with their respective results. Example query considered is "Transferring funds from my savings account to a Fixed Deposit (FD) account, what are the limits and applicable fees?".

In the RAG system, the users are required to manually select the AI service from a range of AI services such as banking transaction services, account details, FD Services, Limits and fee structures. Once the respective service provides the response, the user should corelate the response across multiple services and decide on the next set of actions as shown in Figure 8. This can lead to delays in obtaining a comprehensive response [43].

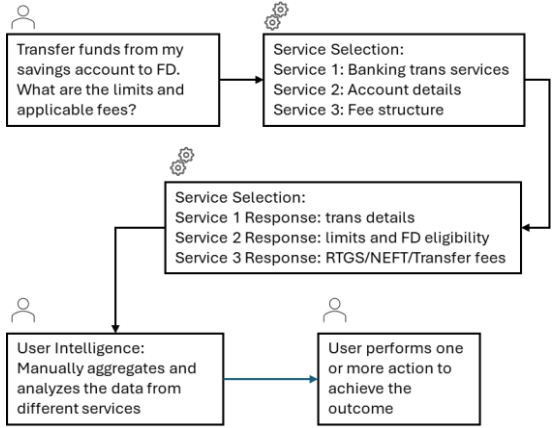

Figure 8: RAG services in product engineering

In MultiFluxAI, for the same query, the orchestration engine automatically determines and invokes the necessary AI services based on the query's context. It dynamically accesses AI services, retrieves account details, FD details, transfer limits, and applicable fees without the user requesting to select or manage specific services as shown in Figure 9. [44,45]. Below are the steps in arriving at the final result for the above query.

**User Prompt P = "Transferring funds from my savings account to a Fixed Deposit, what are the limits and applicable fees?"**

Step 0:

Creation of the Knowledge graph stores (KG) as part of corresponding AI services.

- KG1 = Graph data store, that contains all types of bank accounts and other details offered by the Bank
  {Public Saving account, Corporate Saving account, Corporate Current Account etc.,}
- KG2 = Graph data store, that contains all types of available FDs and their details offered by the Bank
  {Active: 366 days FD – 8.65% interest, 444 days FD – 8.65% interest
  No Active: 500 days FD – 9.0% interest, 270 days FD – 8.25% interest}
- KG3 = Graph data store, that contains all Bank Policy info, Fees, Limits, taxation, breakage clause and others
  {Within bank transfer: fees 1% via RTGS, 1% via NEFT,
  Outside bank transfer: fees 2% via RTGS, 2% via NEFT}

Step 1:

- Sub-Prompt P1 = *Fetch Customer Banking summary*
- Rule1 = Saving account {*KG1 – Public Saving account*}
- CKG1 = Saving Banking account and its details {*Public Saving account*}
- Response 1 (R1) = Customer XXX has greater than ₹*100,000 in his saving account*

Step 2:

- P2 = What are the active FD offered and its details + R1
- Rule2 = FD account
- CKG2 = FD accounts details {Active FDs}

- R2 = There are 2 FDs offered with a minimum deposit amount of ₹100,000 for 366 days and 444 days with an interest rate of 8.65%

Step 3:

- P3 = Banks policy for limits, charges and others + R2
- Rule3 = Policy related to Limits, Fees
- CKG3 = Bank policy {Within bank transfer fees}
- R3 = Charges for transfer of amount from Saving account to FD is 1% for NEFT/RTGS

**Final Result:** *You have sufficient balance for the FD transfer. The daily limit is ₹100,000, with a 1% fee for NEFT/RTGS transfers. Proceed with the transfer, and the applicable charges will be automatically deducted.*

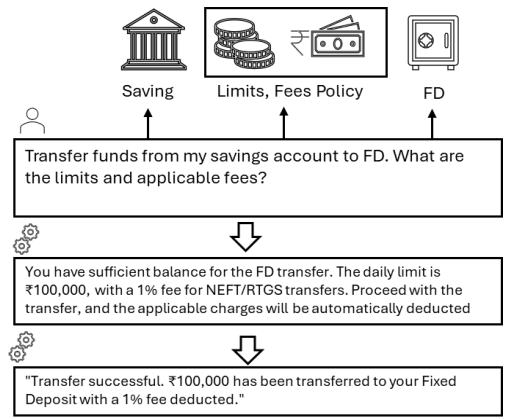

Figure 9: MultiFluxAI intelligence orchestration services

This highlights MultiFluxAI services enhance user experience by leveraging orchestration engine to seamlessly and dynamically access the services based on the query's context. This not only eliminates the need for user intervention but also accelerates the response. Delivers a more integrated, user-friendly experience that significantly improves operational efficiency and user satisfaction.

The observations of the case study are summarized in Table 1.

Table 1: Comparison between Traditional RAG and MultiFluxAI Services

| ID | AI Service Type | Test Description | Steps | Efficiency | Accuracy (%) | Observations |
|---|---|---|---|---|---|---|
| 1 | Traditional RAG | Standard RAG system with real-time retrieval and query processing. | 5-7 steps (manual service selection, data aggregation) | Low (100ms) | 85 | High retrieval latency due to real-time query processing. |
| 2 | MultiFluxAI with Cache | Caching system with preloaded and extended context | 3-4 steps (auto service invocation, data retrieval) | High (20ms) | 92 | Reduced retrieval latency and better accuracy with cached context. |
| 3 | MultiFluxAI with Cache and Rule | Same as Test 2 plus, Business rules and Graph AI | 3-4 steps (auto service invocation, contextual processing) | Very High (0*-10ms) *Knowledge is reused from cache | 95 | Excellent performance with minimal latency and improved accuracy. |

- The MultiFluxAI service outperforms the RAG systems in terms of efficiency, number of steps and accuracy.
- Preloading knowledge and using an extended context store allows the AI service to bypass the retrieval process, leading to efficient responses.
- Caching reduced the response times for queries by over 80%, demonstrating performance improvement when frequently queried data is cached.

The advantages of our approach compared to RAG system:

- Dynamically selects AI services based on the query, reducing user effort and speeding up data retrieval.
- The rule engine ensures relevant, personalized results by applying the right context to each prompt.
- Caching and graph-based management improve response times for frequently queried data.
- Harmonizes data from multiple sources, providing a comprehensive view and eliminating fragmented responses.
- The platform's modular design allows efficient scaling and adapts to evolving needs.

### V. Conclusions

In this paper, we introduced a MultiFluxAI, advanced AI platform designed to overcome the limitations of RAG systems. By integrating intelligent orchestration, caching, rule-based context processing, and graph-based data management. MultiFluxAI addresses the challenges of fragmented data access and manual service selection, thus providing a seamless and dynamic user experience across multiple software products\applications.

Using MultiFluxAI, a case study has been conducted to know the efficiency of the orchestration engine. The metrics show that MutiFluxAI has higher efficiency, accuracy, and minimal number of steps to process different types of user prompts. Future study to be conducted across Retail and Healthcare domains. A detailed study needs to be conducted for the integration of CAG [13].

### VI. References

[1] Alenezi, M., & Akour, M. (2025), "AI-Driven Innovations in Software Engineering: A Review of Current Practices and Future Directions", https://www.mdpi.com/2076-3417/15/3/1344

[2] Crudu, A., & MoldStud Research Team (2024), "The impact of AI and machine learning on software product engineering", https://moldstud.com/articles/p-the-impact-of-ai-and-machine-learning-on-software-product-engineering

[3] Dhaduk, H. (2023), "20+ Product Engineering Challenges Engineers Face (With Possible Solutions)", https://www.simform.com/blog/product-engineering-challenges/

[4] Promact Info Team (2023), "Top Product Engineering Challenges Faced by Teams", https://promactinfo.com/blogs/top-product-engineering-challenges-faced-by-teams/

[5] Upsquare CS Team (2023), "17+ Product Engineering Challenges and Solutions", https://upsquarecs.com/product-engineering-challenges-and-solutions/

[6] MahoutAI Team (2024), "Orchestration Frameworks for Agentic Systems: Strategic Insights Automation", https://mahoutai.com/orchestration-frameworks-for-agentic-systems-strategic-insights-automation/

[7] Lewis, M., Piktus, A., Xu, K., & Stoyanov, V. (2020). Retrieval-Augmented Generation for Knowledge-Intensive NLP Tasks. [https://arxiv.org/abs/2005.11401]

[8] Škorić, M. (2023), "Text vectorization via transformer-based language models and n-gram perplexities", https://arxiv.org/abs/2307.09255

[9] Bursztein, E., Zhang, M., Vallis, O., Jia, X., & Kurakin, A. (2023), "RETVec: Resilient and Efficient Text Vectorizer", https://arxiv.org/pdf/2302.09207v2

[10] Yang, H., Guo, J., Qi, J., Xie, J., Zhang, S., Yang, S., Li, N., & Xu, M. (2024), "A Method for Parsing and Vectorization of Semi-structured Data used in Retrieval Augmented Generation", https://arxiv.org/abs/2405.03989

[11] Ma, X., Zhou, Y., Xu, X., Sun, B., Filev, V., Orlov, N., Fu, Y., & Shi, H. (2022), "Towards Layer-wise Image Vectorization", https://arxiv.org/abs/2206.04655

[12] Dziuba, M., Jarsky, I., Efimova, V., & Filchenkov, A. (2023), "Image Vectorization: a Review", https://arxiv.org/abs/2306.06441

[13] Chao Jin, Zili Zhang, Xuanlin Jiang, Fangyue Liu, Xin Liu, Xuanzhe Liu, Xin Jin. (Apr 2024). RAGCache: Efficient Knowledge Caching for Retrieval-Augmented Generation. [https://arxiv.org/abs/2404.12457]

[14] Yuhan Liu, Hanchen Li, Yihua Cheng, Siddhant Ray, Yuyang Huang, Qizheng Zhang, Kuntai Du, Jiayi Yao, Shan Lu, Ganesh Ananthanarayanan, Michael Maire, Henry Hoffmann, Ari Holtzman, Junchen Jiang, et al. Oct 2023 CacheGen: KV Cache Compression and Streaming for Fast Large Language Model Serving [https://arxiv.org/abs/2310.07240]

[15] Hamza Landolsi, Kais Letaief, Nizar Taghouti, Ines Abdeljaoued-Tej et al. Jan 2025, CAPRAG: A Large Language Model Solution for Customer Service and Automatic Reporting using Vector and Graph Retrieval-Augmented Generation [https://arxiv.org/abs/2501.13993]

[16] Murugan Sankaradas, Ravi K.Rajendran, Srimat T.Chakradhar (Jan 2025) StreamingRAG: Real-time Contextual Retrieval and Generation Framework [https://arxiv.org/abs/2501.14101]

[17] Minaee, S., Kalchbrenner, N., Iandola, F., and Liang, P. (2024), "Large Language Models: A Survey", https://arxiv.org/abs/2402.06196

[18] Naveed, H., Narayanan, P., Ramesh, A., and Fan, J. (2024), "A Comprehensive Overview of Large Language Models", https://arxiv.org/abs/2307.06435

[19] Singh, A., Patel, R., Kumar, V., and Mehta, P. (2025), "Agentic Retrieval-Augmented Generation: A Survey on Agentic RAG", https://arxiv.org/abs/2501.09136

[20] Gupta, S., Rao, K., Desai, M., and Shah, N. (2024), "A Comprehensive Survey of Retrieval-Augmented Generation (RAG): Evolution, Current Landscape and Future Directions", https://arxiv.org/abs/2410.12837

[21] Wang, X., Li, Q., Chen, Z., and Zhang, Y. (2024), "Searching for Best Practices in Retrieval-Augmented Generation", https://arxiv.org/abs/2407.01219

[22] Singh, A., Ehtesham, A., Kumar, S., and Khoei, T. T. (2025), "Agentic Retrieval-Augmented Generation: A Survey on Agentic RAG", https://arxiv.org/abs/2501.09136

[23] Jeong, C., and Jeong, P. (2024), "A Study on the Implementation Method of an Agent-Based Advanced RAG System Using Graph", https://arxiv.org/pdf/2407.19994

[24] Pandya, K., & Holia, M. (2023), "Automating Customer Service using LangChain", https://arxiv.org/pdf/2310.05421

[25] Singh, A., Ehtesham, A., Mahmud, S., & Kim, J.-H. (2024), "Revolutionizing Mental Health Care through LangChain: A Journey with a Large Language Model", https://arxiv.org/abs/2403.05568

[26] Jiao Sun, Q. Vera Liao, Michael Muller, Mayank Agarwal, Stephanie Houde, Kartik Talamadupula, Justin D. Weisz ( Feb 2022),

"Investigating Explainability of Generative AI for Code through Scenario-based Design" https://arxiv.org/abs/2202.04903

[27] Zeqiu Wu, Yushi Hu, Weijia Shi, Nouha Dziri, Alane Suhr, Prithviraj Ammanabrolu, Noah A. Smith, Mari Ostendorf, Hannaneh Hajishirzi, (Jun 2023), "Fine-Grained Human Feedback Gives Better Rewards for Language Model Training", https://arxiv.org/abs/2306.01693

[28] Erik Brynjolfsson, Danielle Li, Lindsey Raymond, (Apr 2023), "Generative AI at Work", https://arxiv.org/abs/2304.11771

[29] Alec Radford, Karthik Narasimhan, Tim Salimans, Ilya Sutskever, "Improving Language Understanding by Generative Pre-Training", https://cdn.openai.com/research-covers/language-unsupervised/language_understanding_paper.pdf

[30] Zihao Li, Zhuoran Yang, Mengdi Wang, (May 2023), "Reinforcement Learning with Human Feedback: Learning Dynamic Choices via Pessimism", https://arxiv.org/abs/2305.18438

[31] Awad, A., Awaysheh, F., & López, H. A. (2025), "BEST: A Unified Business Process Enactment via Streams and Tables for Service Computing", https://arxiv.org/abs/2501.14848

[32] Jeong, S., Baek, J., Cho, S., et al. (2024), "Adaptive-RAG: Learning to Adapt Retrieval-Augmented Large Language Models through Question Complexity", https://arxiv.org/abs/2403.14403

[33] Shi, J., Jain, R., Chi, S., Doh, H., Chi, H., Quinn, A. J., & Ramani, K. (2025), "CARING-AI: Towards Authoring Context-aware Augmented Reality INstruction through Generative Artificial Intelligence", https://arxiv.org/abs/2501.16557

[34] Katrix, R., Carroway, Q., Hawkesbury, R., & Heathfield, M. (2025), "Context-Aware Semantic Recomposition Mechanism for Large Language Models", https://arxiv.org/abs/2501.17386

[35] Siyu Zhou, Tianyi Zhou, Yijun Yang, Guodong Long, Deheng Ye, Jing Jiang, Chengqi Zhang, (Oct 2024), "WALL-E: World Alignment by Rule Learning Improves World Model-based LLM Agents", https://arxiv.org/abs/2410.07484

[36] Shailja Gupta, Rajesh Ranjan, Surya Narayan Singh, (Sept 2024), "Comprehensive Study on Sentiment Analysis: From Rule-based to modern LLM based system", https://arxiv.org/abs/2409.09989

[37] Sajal Regmi, Chetan Phakami Pun, (Nov 2024), "GPT Semantic Cache: Reducing LLM Costs and Latency via Semantic Embedding Caching", https://arxiv.org/abs/2411.05276

[38] Simranjit Singh, Michael Fore, Andreas Karatzas, Chaehong Lee, Yanan Jian, Longfei Shangguan, Fuxun Yu, Iraklis Anagnostopoulos, Dimitrios Stamoulis, (Jun 2024), "LLM-dCache: Improving Tool-Augmented LLMs with GPT-Driven Localized Data Caching", https://arxiv.org/abs/2406.06799

[39] Huachi Zhou, Jiahe Du, Chuang Zhou, Chang Yang, Yilin Xiao, Yuxuan Xie, Xiao Huang, (Jan 2025), "Each Graph is a New Language: Graph Learning with LLMs", https://arxiv.org/abs/2501.11478

[40] Shijie Wang, Wenqi Fan, Yue Feng, Xinyu Ma, Shuaiqiang Wang, Dawei Yin, (Jan 2025), "Knowledge Graph Retrieval-Augmented Generation for LLM-based Recommendation", https://arxiv.org/abs/2501.02226

[41] Sumedh Rasal, (Oct 2024), "A Multi-LLM Orchestration Engine for Personalized, Context-Rich Assistance", https://arxiv.org/abs/2410.10039

[42] Wilson Wei, Nicholas Chen, Yuxuan Li, (Jan 2025), "The Internet of Large Language Models: An Orchestration Framework for LLM Training and Knowledge Exchange Toward Artificial General Intelligence" https://arxiv.org/abs/2501.06471

[43] Amelia Glaese, Nat McAleese, Maja Trebacz, John Aslanides, Vlad Firoiu, Timo Ewalds, Maribeth Rauh, (Sept 2022), "Improving alignment of dialogue agents via targeted human judgements", https://arxiv.org/pdf/2209.14375

[44] Long Ouyang, Jeff Wu, Xu Jiang, Diogo Almeida, Carroll L. Wainwright, Pamela Mishkin, Chong Zhang, Sandhini Agarwal, (May 2022), "Training language models to follow instructions with human feedback", https://arxiv.org/abs/2203.02155

[45] Mary Phuong, Marcus Hutter (Jul 2022), "Formal Algorithms for Transformers", https://arxiv.org/abs/2207.09238